\documentclass[11pt,a4paper]{article}
\usepackage{times,latexsym}
\usepackage{url}
\usepackage[T1]{fontenc}

\usepackage[inline]{enumitem}
\usepackage{hyperref}       
\usepackage{amsfonts}       
\usepackage{nicefrac}       
\usepackage{microtype}      
\usepackage{xcolor}         
\usepackage{booktabs}
\usepackage{amsmath}
\usepackage{multirow}
\usepackage{graphicx}
\usepackage{enumitem}
\usepackage{subcaption}
\usepackage{caption}
\usepackage{rotating}
\usepackage[normalem]{ulem}
\usepackage{amssymb}
\usepackage{colortbl}
\usepackage{linguex}

\usepackage{tikz,pgfplots,pgfplotstable}

\usepackage{siunitx} 
\usepackage{multirow, makecell}
\usepackage{pifont}
\usepackage{siunitx} 
\usepackage{multirow, makecell}
\usepackage{tabularx}
\usepackage[all]{nowidow}
\definecolor{darkgreen}{rgb}{0.0, 0.42, 0.24}
\definecolor{green}{RGB}{112, 173,71}
\definecolor{blue}{RGB}{68, 114,196}
\definecolor{orange}{RGB}{237, 125,49}
\definecolor{red}{RGB}{202, 54,49}
\definecolor{yellow}{RGB}{222,194, 142}

\usepackage[acceptedWithA]{tacl2021v1}

\usepackage{xspace,mfirstuc,tabulary}

\newif\iftaclinstructions
\taclinstructionsfalse 
\iftaclinstructions
\renewcommand{\confidential}{}
\renewcommand{\anonsubtext}{(No author info supplied here, for consistency with
TACL-submission anonymization requirements)}
\newcommand{\instr}
\fi

\iftaclpubformat 

\else

\fi

\title{Is Context Helpful for Chat Translation Evaluation?}

\author{Sweta Agrawal$^{1}$\thanks{\nobreak\hspace{.16667em plus .08333em}
 Work partially developed during internship at Unbabel.}, Amin Farajian$^{2}$, Patrick Fernandes$^{1,4,5}$ \\
 \textbf{Ricardo Rei$^{1,2,3}$, André F.T. Martins$^{1,2,4}$} \\
  $^1$Instituto de Telecomunicações, $^2$Unbabel, $^3$INESC-ID \\
  $^4$Instituto Superior Técnico \& Universidade de Lisboa (Lisbon ELLIS Unit) \\
  $^5$Carnegie Mellon University \\
  \texttt{swetaagrawal20@gmail.com} \\
  \\}

\date{}

\begin{document}
\maketitle
\begin{abstract}
Despite the recent success of automatic metrics for assessing translation quality, their application in evaluating the quality of machine-translated chats has been limited. Unlike more structured texts like news, chat conversations are often unstructured, short, and heavily reliant on contextual information. This poses questions about the reliability of existing sentence-level metrics in this domain as well as the role of context in assessing the translation quality. Motivated by this, we conduct a meta-evaluation of existing sentence-level automatic metrics, primarily designed for structured domains such as news, to assess the quality of machine-translated chats. We find that reference-free metrics lag behind reference-based ones, especially when evaluating translation quality in out-of-English settings. We then investigate how incorporating conversational contextual information in these metrics affects their performance. Our findings show that augmenting neural learned metrics with contextual information helps improve correlation with human judgments in the reference-free scenario and when evaluating translations in out-of-English settings.  Finally, we propose a new evaluation metric, \textsc{Context-MQM}, that utilizes bilingual context with a large language model (LLM) and further validate that adding context helps even for LLM-based evaluation metrics.

\end{abstract}


\section{Introduction}

Automatically estimating the quality of machine or human-generated translations has received a lot of attention over the past two decades from the NLP community \cite{han-etal-2021-translation}, specifically via shared tasks organized by WMT since 2014\--present \cite{machacek-bojar-2014-results, stanojevic-etal-2015-results, bojar-etal-2016-results, bojar-etal-2017-results, ma-etal-2018-results, ma-etal-2019-results, mathur-etal-2020-results, freitag-etal-2021-results, freitag-etal-2022-results, freitag-etal-2023-results}. A variety of evaluation metrics have been developed for this purpose, encompassing lexical matching approaches such as BLEU \cite{papineni2002bleu}, \textsc{chrF} \cite{popovic2015chrf}, and METEOR \cite{lavie-agarwal-2007-meteor}; embedding-based methods like \textsc{BERTScore} \cite{zhang2019bertscore} and Word Mover Distance \cite{zhao-etal-2019-moverscore}; learned metrics like \textsc{Comet} \cite{rei-etal-2020-comet}, and BLEURT \cite{sellam2020bleurt}; and metrics that employ prompting techniques with large language models (LLMs) like \textsc{Gemba-MQM} \cite{kocmi-federmann-2023-gemba} or \textsc{AutoMQM} \cite{fernandes-etal-2023-devil}.

Among these metrics, neural learned metrics have gained widespread acceptance \cite{freitag-etal-2022-results} as they are directly trained to predict sentence-level translation quality assessment scores \cite{kreutzer-etal-2015-quality, rei-etal-2020-comet, sellam2020bleurt}, word-level error annotations collected by professional linguists \cite{guerreiro2023xcomet} or post-editing efforts as measured by \textsc{HTER} \cite{snover-etal-2006-study, fonseca-etal-2019-findings, specia-etal-2021-findings}.
However, the reliance on human-written reference translations and judgments collected predominantly from structured domains like news or Wikipedia as training data raises questions about their adaptability and reliability in detecting errors in other domains \cite{zouhar2024finetuned}, for example in evaluating the quality of translations in more informal settings. 

\begin{figure}[t]
\centering
\begin{subfigure}{\linewidth}
  \centering
\includegraphics[width=0.8\textwidth]{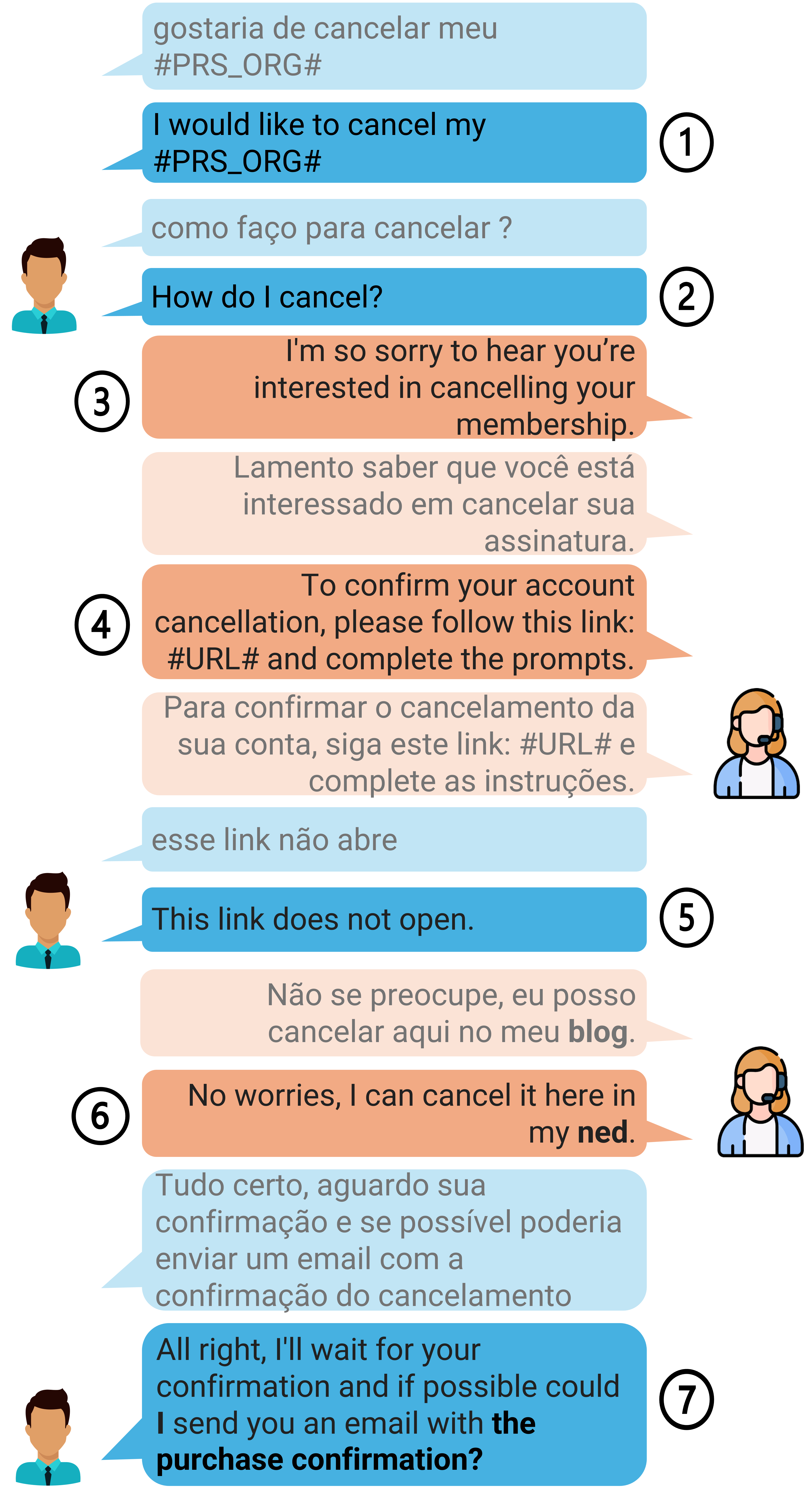}
\end{subfigure}%
 \caption{An example bilingual conversation from the MAIA corpus \cite{martins-etal-2020-project}: the agent and the customer only see the texts in English and Portuguese, respectively. The errors (both MT and user-generated) are bold-faced. 
 }
 \vspace{-0.5cm}
 \label{fig:chat_example}
\end{figure}

Unlike news articles which involve carefully authored and well-formatted text, which current translation systems are well equipped for, chat conversations are often synchronous and short, and involve formal language, colloquial expressions, and slang that may not have direct equivalents in the target language \cite{goncalves-etal-2022-agent}. An example of such a conversation is presented in Figure~\ref{fig:chat_example}. Although chat messages are generally brief and easier to translate, errors introduced by machine translation (MT) systems might go unnoticed by end users if not detected properly, potentially leading to miscommunication, conversation breakdowns, or even more serious implications on high-risk domains (e.g. patient-physician chats) \cite{yamashita2009difficulties, robertson2022understanding}. 

Moreover, conversational texts rely heavily on context, meaning that the interpretation of a text is largely influenced by the surrounding contextual information. Hence, MT systems for such domains are often trained with contextual information and this has been shown to improve translation quality, lexical inconsistency, and coherence of the generated outputs \cite{farinha-etal-2022-findings, fernandes-etal-2021-measuring}. However, it is unclear to what extent context plays a role in estimating translation quality for machine-translated conversations.  

In this work, we first systematically analyze the nature and the frequency of the errors in real bilingual chat translations from customer support and contrast them with the structured news domain (\S~\ref{sec:newsvsconversation}). We find that translation errors are 21\% less frequent in chat relative to the news domain and that the nature of the errors introduced by MT systems in the two domains is also different. This underscores the importance of understanding and evaluating existing automatic metrics for chat translation. Due to the infrequent error occurrences at the sentence level, MT systems for chat translation tasks might receive higher scores, necessitating a more nuanced evaluation of the metrics. 

Motivated by the above, we present a meta-evaluation of existing sentence-level automatic metrics, primarily tested on news translation tasks, in their ability to gauge the quality of machine-translated chats (\S~\ref{sec:metaeval}). We evaluate these metrics across two scenarios: on \textit{all} translation pairs as well as \textit{imperfect} translations as judged by humans. We then study the impact of augmenting a subset of these learned neural metrics with different types of conversational contextual information (\S~\ref{sec:chatcontext}). Our findings are summarized below:

\begin{itemize}
\itemsep0em 
    \item  \textsc{Comet-22}, a reference-based metric, achieves the highest overall correlation with human judgments.
    \item Reference-based \textsc{Comet-22} does not benefit from the added contextual information, whereas, reference-free \textsc{Comet-20-QE} has better correlations with human judgments as the context window increases when evaluating translations in out-of-English settings. This is useful as reference-free metrics allow translation quality to be assessed on the fly.
    \item Adding context helps assess translation quality better for short and ambiguous sentences. Using correct and complete context from both speakers is essential for improving chat quality estimation via neural learned metrics.
\end{itemize}

Finally, we present \textsc{Context-MQM}, an LLM-based evaluation metric that utilizes context for chat translation quality estimation (\S~\ref{sec:contextmqm}). Our preliminary experiments with \textsc{Context-MQM} show that adding bilingual context to the evaluation prompt helps improve the quality assessment of machine-translated chats on imperfect translations.


\section{Errors in Chat vs. News: A Case Study} \label{sec:newsvsconversation}



To better understand how the error types differ in these domains, we present an analysis of the nature and the frequency of errors using Multidimensional Quality Metrics (MQM) annotations (conversation: 7120, news: 4800 translation pairs) 
collected for English-German as a part of the WMT22 Metrics shared task \cite{freitag-etal-2022-results}. 

\begin{figure}[t]
\centering
\begin{subfigure}{0.75\linewidth}
  \centering
\includegraphics[width=\textwidth]{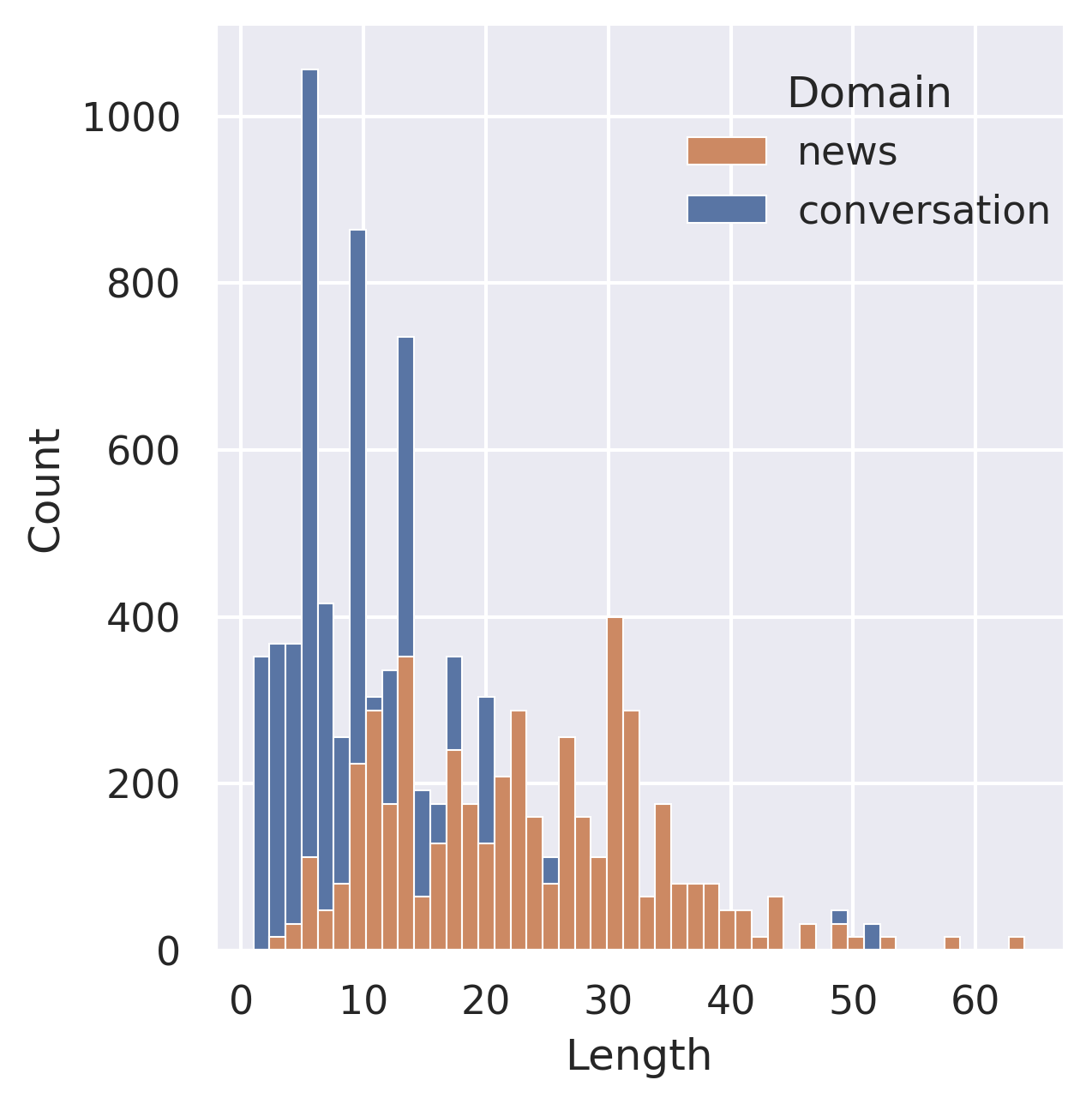}
\end{subfigure}%
 \caption{Conversational texts tend to be much shorter relative to news texts in the WMT22 English-German dataset.
 }
 \label{fig:length}
 \vspace{-0.5cm}
\end{figure}


\paragraph{Errors are less frequent.} We calculate the percentage of translations with a perfect MQM score from the news and conversational subset of the WMT22 dataset: only 46.4\% of news translations have perfect MQM scores, whereas this percentage is 57.8\% for the conversational domain. This suggests that errors are less frequent in the conversational domain compared to the news, likely due to the relatively short (see Figure~\ref{fig:length}) and probably less complex text dominant in conversations. However, it is important to note that these errors do not occur in isolation and have the potential to escalate into larger issues, leading to significant breakdowns in communication.  Moreover, as the errors are less frequent, the true quality of the model-generated translations could be misrepresented if they are not accurately identified by quality estimation systems. 

\begin{figure}[t]
\centering
  \includegraphics[width=\linewidth]{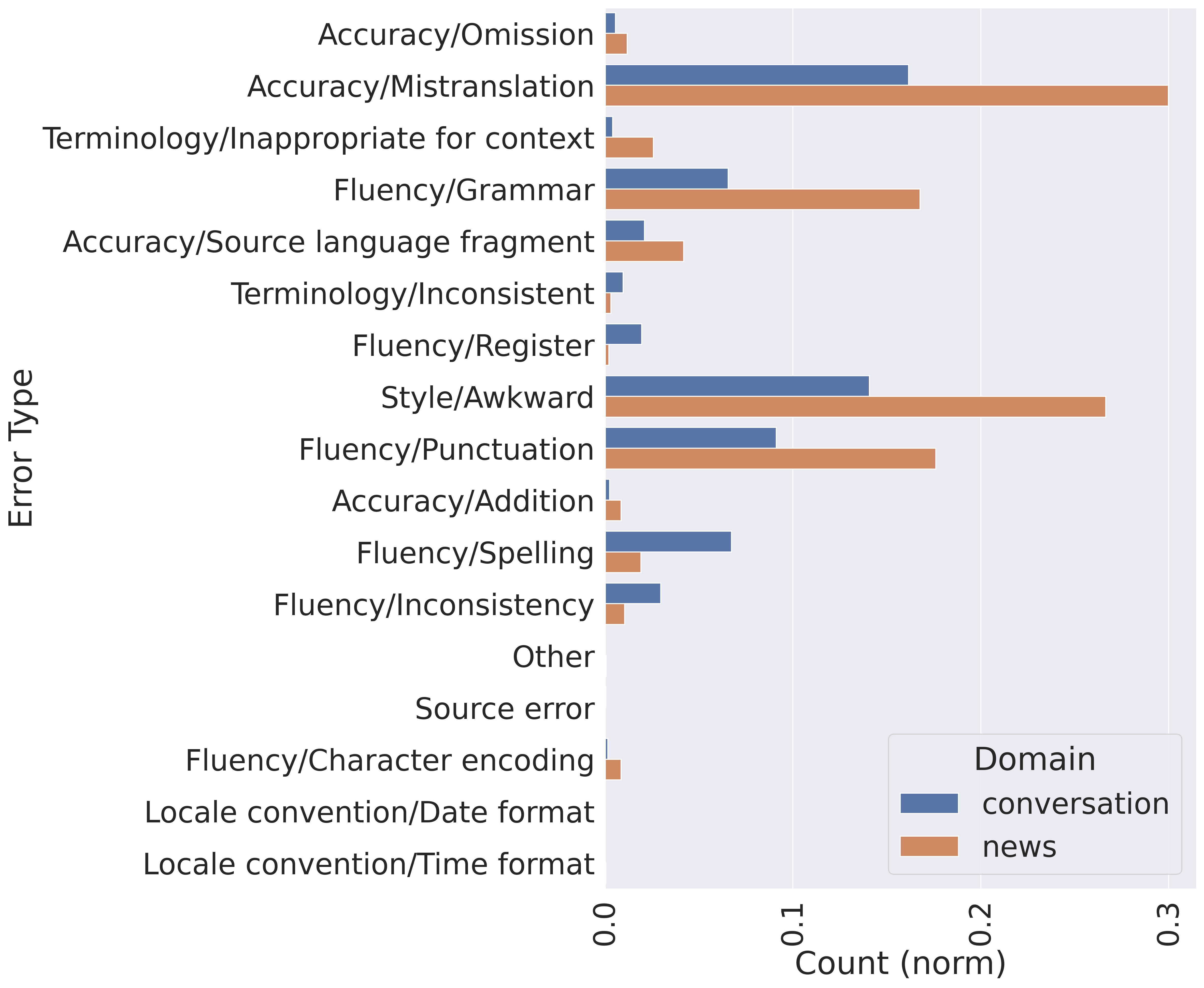}
    \caption{Counts of MQM error categories normalized by the number of annotated instances for each domain: frequent errors differ in the two domains.  
    }  \label{fig:error_dist_news_conv}
     \vspace{-0.5cm}
\end{figure}

\paragraph{Most frequent error types across domains are different.} Errors related to fluency, such as issues with spelling, consistency, and register, occur more frequently in conversations compared to accuracy-based errors like mistranslation, which are more common in the news domain (Figure~\ref{fig:error_dist_news_conv}). This underscores the idea that the nature of observed errors is also influenced by the specific domain context.

The variation in error types and their frequencies across the two domains underscores the need for a systematic study of the suitability of current automatic metrics within the chat domain.

\section{Meta Evaluation of Automatic Metrics} \label{sec:metaeval}

We now turn towards investigating to what extent automatic sentence-level MT evaluation metrics frequently used in the literature capture the translation quality of conversational data. We detail the setup used for our meta-evaluation below:

\begin{table*}[ht]
\centering
 \setlength\tabcolsep{2.3pt}
\scalebox{0.78}{
\begin{tabular}{llcccrrrrrrrrr}
 \toprule
\multirow{2}{*}{\textbf{LP}}& \multirow{2}{*}{\textbf{\textsc{Sender}}} & \multicolumn{3}{c}{\textsc{\textbf{Count}}} & \multicolumn{3}{c}{\textsc{\textbf{Length}}} & \multicolumn{4}{c}{\textbf{CUA}}& \multirow{2}{*}{\textbf{\% Perfect MQM}}  \\
 &  & \# (Instances) & \# (Chats)  & \# (Systems) & Source & MT & Ref & Weak & Moderate & Good & Excellent   \\
\midrule
\textsc{En-De} & Agent & 2715 & 30 & 5 & 48.82 & 57.34 & 57.20 & 89 & 90 & 120 & 2219 & 73.6  \\
\textsc{De-En} & Customer & 2720 &  &  & 38.14 & 34.46 & 35.49 & 113 & 88 & 166 & 2067 & 65.4 \\
\textsc{En-Fr} & Agent & 1868 & 23 & 2 & 36.26 & 42.81 & 44.21 & 100 & 118 & 129 & 1274 & 46.7 \\
\textsc{Fr-En} & Customer & 1020 & & & 35.72 & 32.36 & 33.19 & 45 & 37 & 41 & 748 & 65.7 \\
\textsc{En-Pt} & Agent & 1318 & 28 & 2 & 43.21 & 46.34 & 45.96 & 77 & 81 & 103 & 872 & 50.1 \\
\textsc{Pt-En} & Customer & 1016 & & & 30.44 & 31.06 & 31.35 & 40 & 68 & 66 & 624 & 54.5 \\
  \bottomrule
 \end{tabular}
 }
\caption{Statistics extracted from the human annotations of the WMT 22 Chat Shared Task: Errors are less frequent in the chat domain with 46.7\% to 73.6\% translations receiving perfect MQM scores.}\label{tab:data_stats} 
\end{table*}

\subsection{Dataset} 

We use the MQM annotations collected from the WMT 2022 Chat Shared Task \cite{farinha-etal-2022-findings}. The dataset consists of genuine bilingual customer support conversations which were translated by participants' submitted automatic MT systems. The outputs from the submitted MT systems were then evaluated via an MQM-based human evaluation that is tailored to assess customer support content. The evaluation was conducted by Unbabel's in-house professional linguists and translators using the whole conversational context.

We convert token-level MQM spans into a sentence-level score via the following formula:
\begin{equation}
    \mathrm{MQM} = - (C_\mathrm{Min} + 5 \times C_\mathrm{Maj} + 10 \times C_\mathrm{Cri})
\end{equation}
where, $C_\mathrm{Min}$, $C_\mathrm{Maj}$ and $C_\mathrm{Cri}$ denote the number of minor, major, and critical errors respectively \cite{lommel2014multidimensional, farinha-etal-2022-findings}. 

Table~\ref{tab:data_stats} shows several statistics extracted from the data.\footnote{Note that the Agent communicates always in English and the Customer in non-English languages.} Across different language pairs, the percentage of instances with no errors, i.e., \% Perfect MQM ranges from  46.7 to 73.6\%, which further confirms our initial analysis that errors are less frequent in translated chats. Besides the 
\% perfect MQM, following \citet{farinha-etal-2022-findings}, we also present  Customer Utility Analysis (CUA) which buckets MQM scores in four regions: Weak (negative - 39 MQM); Moderate (40 - 59 MQM); Good (60 - 79 MQM) and Excellent (80 - 100 MQM). From the table, it is apparent that, overall, translating agent directions tends to be easier compared to translating customer directions as indicated by the higher counts in the ``good $+$ excellent'' categories, a finding highlighted by \citet{farinha-etal-2022-findings}.

\subsection{Metrics}

We benchmark several metrics spanning lexical, embedding-based, and neural metrics frequently used for translation evaluation: 

\paragraph{BLEU} \cite{papineni2002bleu} estimates the translation quality based on n-gram overlap between the hypotheses and references. We compute sentence-level BLEU \cite{chen-cherry-2014-systematic} using the \textsc{Sacrebleu} \cite{post-2018-call} library.\footnote{\url{https://github.com/mjpost/sacrebleu/}: nrefs:1$\vert$case:mixed$\vert$eff:no$\vert$tok:13a$\vert$smooth:exp$\vert$version:2.4.0}
\paragraph{\textbf{\textsc{chrF}}} \cite{popovic2015chrf} evaluates the similarity by computing an F1 score between the overlapping character n-grams in the hypotheses and references.
\paragraph{\textsc{BERTScore}} \cite{zhang2019bertscore} uses the cosine similarity between the pre-trained contextualized embeddings to measure the similarity between the hypotheses and references.
\paragraph{\textbf{BLEURT}} \cite{sellam2020bleurt} utilizes pre-trained transformer models to estimate a quality score that reflects the semantic similarity between the hypothesis and the reference. BLEURT is a neural regression metric trained on an existing collection of human judgments. 
\paragraph{\textsc{Comet-22}} \cite{rei-etal-2022-comet} is an XLM-R-based \cite{conneau2020unsupervised} regression model trained on direct assessments from WMT17 to WMT20 and provides scores ranging from 0 to 1. Unlike BLEURT which only utilizes the reference and the hypothesis texts, COMET uses the triplet (source, hypothesis, reference) to return a score that reflects the quality of the hypothesis relative to both source and reference.
\paragraph{\textbf{\textsc{Comet-20-QE}}} \cite{rei-etal-2020-unbabels} is also an XLM-R-based regression metric trained to predict the direct assessment scores using representations extracted from the (source, hypothesis) pair.
\paragraph{\textbf{\textsc{CometKiwi-22-QE}}} \cite{rei-etal-2022-cometkiwi} is a reference-free model built on top of the \textsc{InfoXLM-R} model \cite{chi-etal-2021-infoxlm} and is trained to predict direct assessments from WMT17-20 and the MLQE-PE corpus \cite{fomicheva-etal-2022-mlqe}. It generates scores ranging from 0 to 1. 
\paragraph{\textbf{\textsc{xComet-XL}}} \cite{guerreiro2023xcomet}, with 3.5B parameters, is an explainable learned metric trained to predict both sentence-level quality scores and MQM-like error spans from the (source, hypothesis, reference) triplet. Furthermore, the metric can also be used for quality estimation by using only the source and the hypothesis as input, referred to as \textsc{xComet-QE-XL}.
\paragraph{\textbf{\textsc{MetricX-23-XL}}} \cite{juraska-etal-2023-metricx} is a regression-based metric built on top of mT5 \cite{xue-etal-2021-mt5} and is trained to regress the true MQM score to predict an error score in the range [0, 25]. The input to the model is the concatenated hypothesis and reference translations. \\
\paragraph{\textbf{\textsc{MetricX-23-QE-XL}}} \cite{juraska-etal-2023-metricx} is a reference-free version of \textsc{MetricX-23-XL} that instead takes as input the source and the hypothesis.  \\

Following the WMT QE shared task evaluation \cite{blain-etal-2023-findings}, we report Spearman-rank 
correlation \cite{zar2005spearman} to assess how well the automatic metric scores correlate with human judgments.\footnote{While the QE shared task uses Spearman rank, the Metrics shared task employs both Pearson and Kendall Tau. In this work, we opted for Spearman since it serves as a compromise between Pearson and Kendall \cite{deutsch-etal-2023-ties}.}


\begin{table*}[ht]
\centering
\scalebox{0.80}{
\begin{tabular}{l@{\hskip 0.2in}lrrrrrrrr}
 \toprule
 & \multirow{2}{*}{\textsc{\textbf{Metric}}} & \multicolumn{2}{c}{\textsc{\textbf{Average}}}  & \multicolumn{2}{c}{\textsc{\textbf{En-De}}} & \multicolumn{2}{c}{\textsc{\textbf{En-Fr}}} & \multicolumn{2}{c}{\textsc{\textbf{En-Pt}}} \\
 &  & All & Imperfect  & All & Imperfect & All & Imperfect  & All & Imperfect \\
\midrule
\multirow{7}{*}{\rotatebox[origin=c]{90}{\textsc{Ref-based}}} & \textsc{chrF} & 0.531 &0.404 &0.430 &0.253 &0.666 &0.510 &0.496 &0.448 \\
& \textsc{BLEU} & 0.500 &0.319 &0.363 &0.151 &0.642 &0.426 &0.494 &0.381\\
& \textsc{BERTScore} & 0.545 &0.403 &0.439 &0.250 &0.680 &0.501 &0.516 &0.457 \\
& \textsc{BLEURT} &0.372 &0.511 &0.398 &0.411 &0.326 &0.663 &0.393 &0.458\\
& \textsc{Comet-22} &  \textbf{0.633} & \textbf{0.551} & \textbf{0.578} & \textbf{0.445} & 0.721 &0.634 &0.602 & \textbf{0.573}\\
& \textsc{xComet-XL} &0.402 &0.525 &0.402 &0.432 &0.381 &0.649 &0.422 &0.494 \\
& \textsc{MetricX-23-XL} & 0.622 & 0.543 & 0.535 & 0.404 & \textbf{0.723} & \textbf{0.692} & \textbf{0.609} & 0.535 \\
\addlinespace[0.2cm]

\multirow{4}{*}{\rotatebox[origin=c]{90}{\textsc{Ref-Free}}} & \textsc{Comet-20-QE} &0.379 &0.316 &0.410 &0.294 &0.376 &0.339 &0.351 &0.315 \\
& \textsc{CometKiwi-22-QE} & 0.300 & \textbf{0.416} &0.366 &0.416 &0.203 & \textbf{0.508} &0.331 &0.325\\
& \textsc{xComet-QE-XL} & 0.276 &0.363 &0.349 &0.376 &0.160 &0.383 &0.318 &0.329\\
& \textsc{MetricX-23-QE-XL} & \textbf{0.447} &0.402 & \textbf{0.438} & \textbf{0.371} & \textbf{0.437} &0.479 & \textbf{0.466} & \textbf{0.356} \\
  \bottomrule
 \end{tabular}
 }
\caption{\textsc{Comet-22} achieves the highest correlation on average across all \textbf{Agent} language pairs with \textsc{MetricX-23-XL} as a close competitor in the \textsc{Ref-Based} setup. \textsc{MetricX-23-QE-XL} outperforms \textsc{Comet} alternatives in the \textsc{ref-Free} setting.}\label{tab:benchmarkoutofeng} 
\end{table*}

\begin{table*}[ht]
\centering
\scalebox{0.80}{
\begin{tabular}{l@{\hskip 0.2in}lrrrrrrrr}
 \toprule
 &\multirow{2}{*}{\textsc{\textbf{Metric}}} & \multicolumn{2}{c}{\textsc{\textbf{Average}}}  & \multicolumn{2}{c}{\textsc{\textbf{De-En}}} & \multicolumn{2}{c}{\textsc{\textbf{Fr-En}}} & \multicolumn{2}{c}{\textsc{\textbf{Pt-En}}} \\
 & & All & Imperfect  & All & Imperfect  & All & Imperfect  & All & Imperfect  \\
\midrule
\multirow{7}{*}{\rotatebox[origin=c]{90}{\textsc{Ref-Based}}} &\textsc{chrF} & 0.427 &0.188 &0.400 &0.201 &0.411 &0.157 &0.469 &0.205 \\
 &\textsc{BLEU} & 0.396 &0.166 &0.390 &0.154 &0.373 &0.147 &0.425 &0.198 \\
 &\textsc{BERTScore} & 0.484 &0.280 &0.445 &0.239 &0.467 &0.332 &0.539 &0.269  \\
 &\textsc{BLEURT} & 0.559 &0.451 &0.540 &0.445 &0.520 &0.464 &0.617 &0.444 \\
 &\textsc{Comet-22} & \textbf{0.610} &0.517 & 0.580 &0.438 &\textbf{0.588} &0.535 &\textbf{0.661} &0.578 \\
 &\textsc{xCOMET-XL} & 0.454 & \textbf{0.566} &0.357 &\textbf{0.514} &0.479 &\textbf{0.588} &0.527 &0.594 \\
 &\textsc{MetricX-23-XL} & 0.608 &	0.551 &	\textbf{0.589} &	0.511 &	0.583 & 0.544& 0.651 &\textbf{0.598} \\

\addlinespace[0.2cm]

 \multirow{4}{*}{\rotatebox[origin=c]{90}{\textsc{Ref-Free}}}&\textsc{Comet-20-QE} & \textbf{0.516} &0.415 &\textbf{0.554} &0.381 &0.471 &0.429 &0.523 &0.435 \\
 &\textsc{CometKiwi-22-QE} &0.438 &0.443 &0.385 &0.463 &0.456 &0.406 &0.473 &0.461 \\
 &\textsc{xCOMET-QE-XL} &0.447 &0.493 &\textbf{0.388} &\textbf{0.492} &0.462 &\textbf{0.479} &\textbf{0.492} &0.508 \\
 &\textsc{MetricX-23-QE-XL} & 0.395 & \textbf{0.497} &0.383 &0.490 &0.382 &0.431 &0.420 &\textbf{0.569} \\
  \bottomrule
 \end{tabular}
 }
\caption{On average, on the \textbf{Customer} language pairs,  \textsc{Comet-22} and \textsc{MetricX-23-XL} achieve the highest correlation scores and neural learned metrics consistently outperform lexical metrics.}\label{tab:benchmarkintoeng} 
\vspace{-0.4cm}
\end{table*}

\section{Context-Aware Translation Evaluation } \label{sec:chatcontext}


Let Agent (A) and Customer (C) represent the two participants in a bilingual chat, where A and C communicate in languages $l_a$ and $l_c$ respectively. Given a text $x$ generated by A or C, the goal is to predict the quality of its translation, $\hat{y}$. We extend existing sentence-level automatic metrics to utilize conversational context as detailed below:

\paragraph{Incorporating Context} Following \citet{vernikos-etal-2022-embarrassingly}, for a neural learned metric like \textsc{Comet}, we obtain the contextual representations of the current source, $x$, and the reference ($y$)/hypothesis ($\hat{y}$) sentences by prepending up to $k$ sentences of context preceding to it. However, only the representations of the current instance are pooled before passing them to the regressor module that generates the quality score. This allows the learned metric to be informed by the surrounding context when evaluating the translation quality of the current instance. This simple approach was used to evaluate document-level translation quality and was shown to be more effective over sentence-level counterparts \cite{vernikos-etal-2022-embarrassingly}, whereas, in our work, we study and extend its applicability to contextualized sentence-level chat quality estimation. We emphasize that our goal is not to evaluate translation quality at the conversation level but rather to utilize context to inform the sentence-level assessment, which renders  document-level metrics that aggregate scores over sentences like \textsc{SliDe} \cite{raunak-etal-2023-evaluating} 
not applicable to our use case.

\paragraph{Choice of Context} We explore the usage of two types of contextual information for translation quality estimation: \textit{within} and \textit{across} participants. The current text, assuming it is generated by Customer C, can be preceded by the context from previous turns by the same participant (\textit{within}), or the context from both participants (\textit{across}). For example, for the source text generated by Customer C at time $t=7$ in Figure~\ref{fig:chat_example}, the two preceding contextual sentences ($k=2$) for both settings are shown: 
\begin{quotation}
\small
\noindent
 \underline{Context (\textit{within})}: \\
$t=2$ \\
Original: {\fontfamily{cmss}\selectfont como faço para cancelar ?}\\
Translation: {\fontfamily{cmss}\selectfont How do I cancel?}\\
$t=5$ \\
Original: {\fontfamily{cmss}\selectfont esse link não abre}\\
Translation: {\fontfamily{cmss}\selectfont This link does not open.}\\

\noindent
\underline{Context (\textit{across}):} \\
$t=5$ \\
Original: {\fontfamily{cmss}\selectfont esse link não abre}\\
Translation: {\fontfamily{cmss}\selectfont This link does not open.}\\
$t=6$ \\
Original: {\fontfamily{cmss}\selectfont No worries, I can cancel it here in my ned.} \\
Translation: {\fontfamily{cmss}\selectfont Não se preocupe, eu posso cancelar aqui no meu blog.}
\end{quotation}

Adding context in the \textit{within} setting is straightforward, we prepend up to $k$ sentences of source and translation context to the current source ($x$) and the hypothesis ($\hat{y}$)/reference ($y$) from the \textit{same} participant separated by a tag, \texttt{$<$sep$>$}. For the \textit{across} setting, to have the context in the same language on each side, we also prepend the translated context to the source ($x$) and the source context to the hypothesis ($\hat{y}$)/reference ($y$) from the \textit{other} participant. Using our proposed extensions of the sentence-level metrics, we study how they impact the evaluation of translation quality.  





\section{Results}

We first present the results of the meta-evaluation of existing automatic sentence-level MT metrics. Then we show the impact of adding context to a subset of these metrics with ablations on how context impacts translation quality evaluation. Finally, we present a preliminary study on utilizing context with LLM-based MT evaluation.

\subsection{Meta Evaluation}

Tables ~\ref{tab:benchmarkoutofeng} and ~\ref{tab:benchmarkintoeng} show the correlation of human judgments with automatic metrics for all the Agent and the Customer language pairs respectively. ``Imperfect'' translations are  instances marked with an MQM score of $<$0 by humans. 

Overall, in the reference-based setup, when considering all the instances in the corpus (``All''), \textsc{Comet-22} achieves the highest Spearman Correlation on both settings (Agent and Customer), outperforming all other metrics, with \textsc{MetricX-23-XL} as a close competitor. For reference-free evaluation, \textsc{MetricX-23-QE-XL} and \textsc{Comet-20-QE} achieve the highest correlation with human judgments when evaluating translations out of English and into English respectively. However, there is a big gap between the best-performing metric for the reference-free and reference-based evaluation via learned metrics on average across all Agent ($\delta$(\textsc{Comet-22},\textsc{MetricX-23-QE-XL}): $0.186$) and Customer ($\delta$(\textsc{Comet-22},\textsc{Comet-20-QE}): $0.094$) language pairs 
in the chat domain.

When considering the imperfect translations, there is no clear winner: most metrics suffer a drop in correlation for this subset compared to ``All'' translations except \textsc{xComet-XL} and \textsc{xComet-QE-XL}. \textsc{xComet-XL} consistently achieves better correlations on this subset than ``ALL'' data. We hypothesize that this could be due to an over-prediction of errors via the metric for chats. 

\begin{figure*}[t]
 \centering
\begin{subfigure}{0.30\textwidth}
  \centering
  \includegraphics[width=\linewidth, trim={0 0 15cm 0},clip]{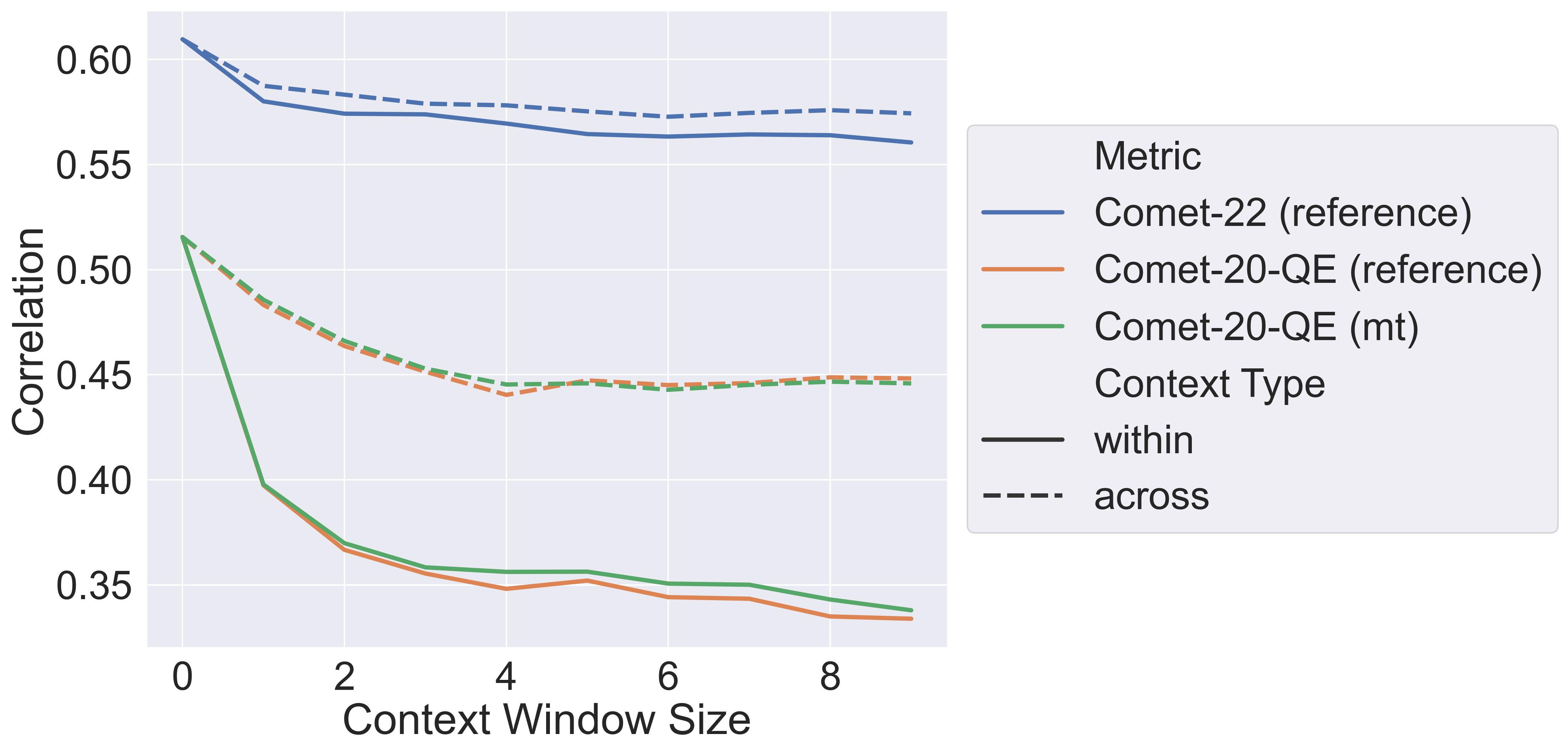}
   \caption{Into-English (Customer)}
\end{subfigure}%
\begin{subfigure}{0.48\textwidth}
  \centering
  \includegraphics[width=\linewidth]{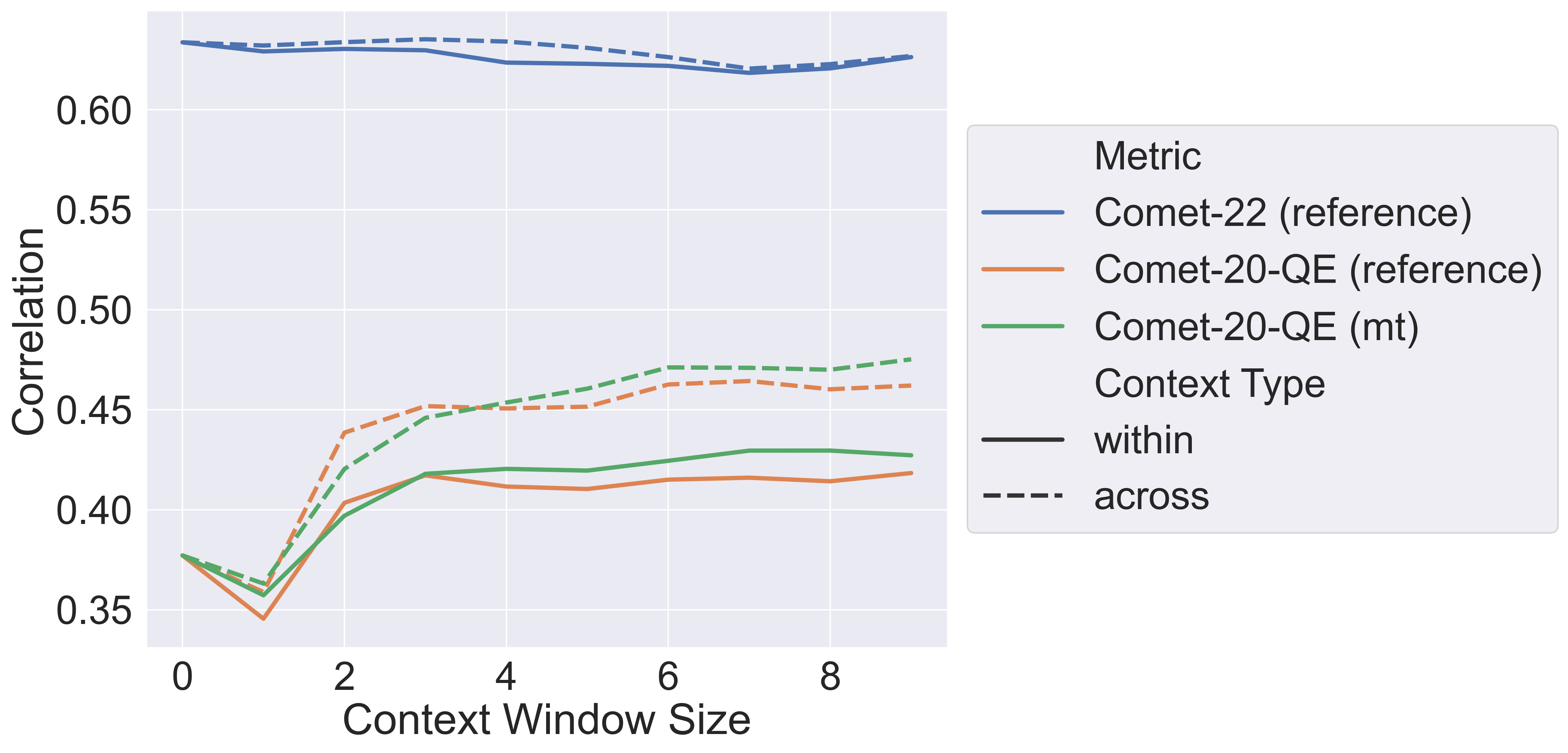}
   \caption{Out-of-English (Agent)}
\end{subfigure}%
    \caption{ Impact of varying context window and context type (across/within) on average correlation across Agent and Customer settings: adding complete context (across) helps improve metrics performance in out-of-English reference-free settings (Agent) but is detrimental for into-English (Customer) evaluation. 
    }  \label{fig:context}
\end{figure*}

\begin{figure*}[htb!]
\begin{subfigure}{0.25\textwidth}
\centering
  \includegraphics[width=\linewidth]{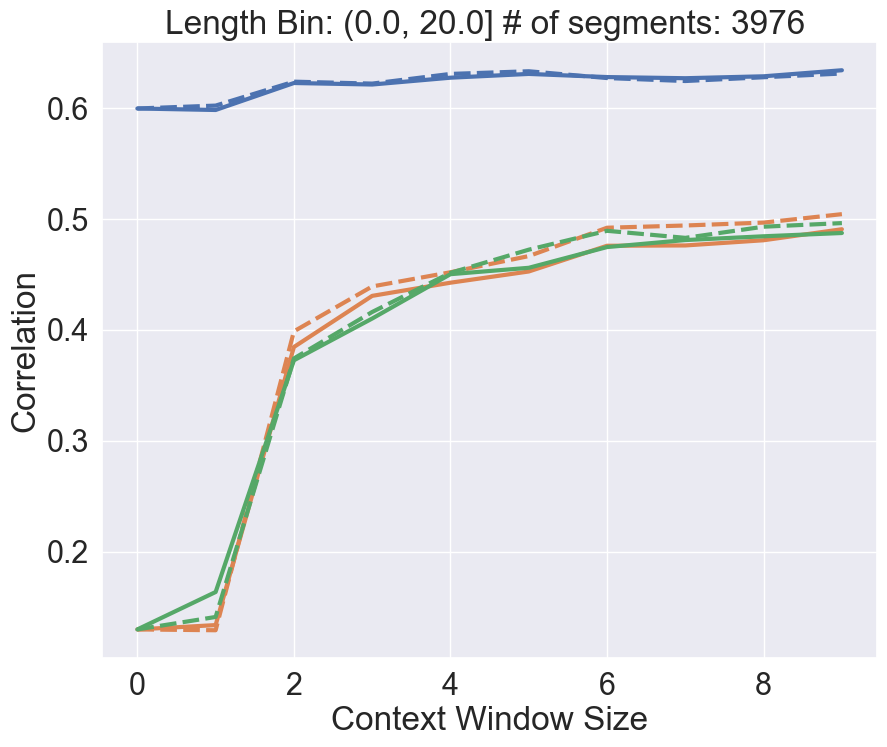}
\end{subfigure}
\begin{subfigure}{0.25\textwidth}
  \centering
  \includegraphics[width=\linewidth]{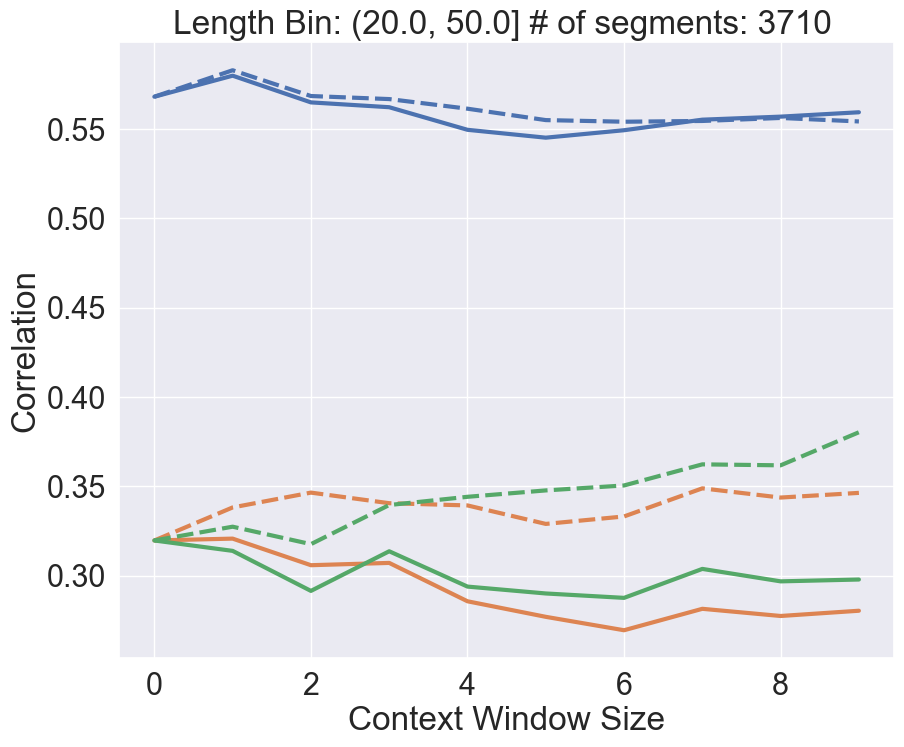}
\end{subfigure}%
\begin{subfigure}{0.25\textwidth}
  \centering
  \includegraphics[width=\linewidth]{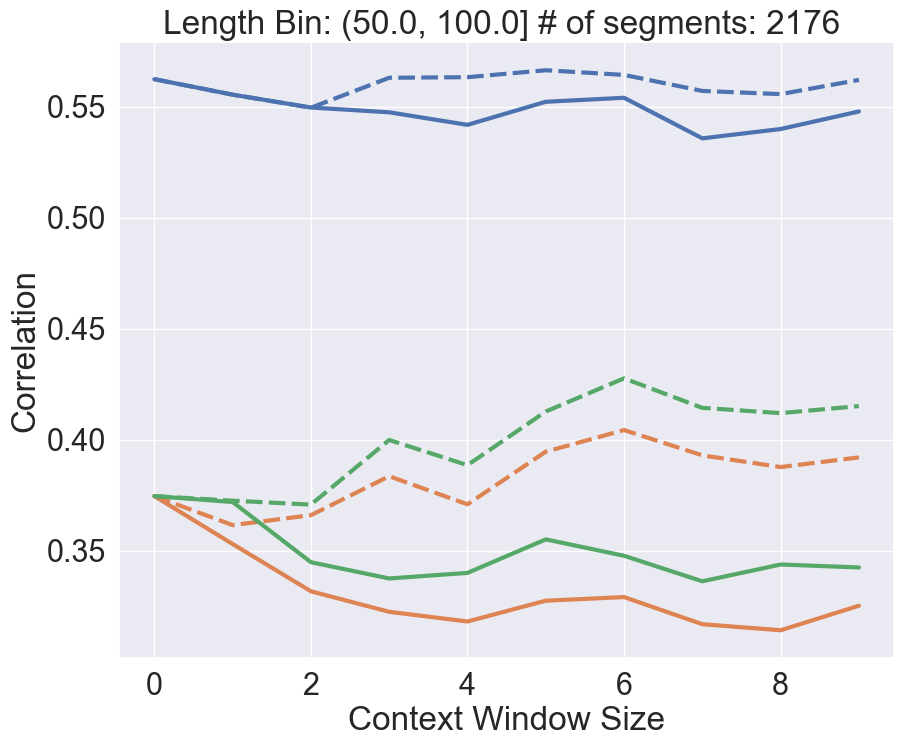}
\end{subfigure}%
\begin{subfigure}{0.25\textwidth}
  \centering
  \includegraphics[width=\linewidth]{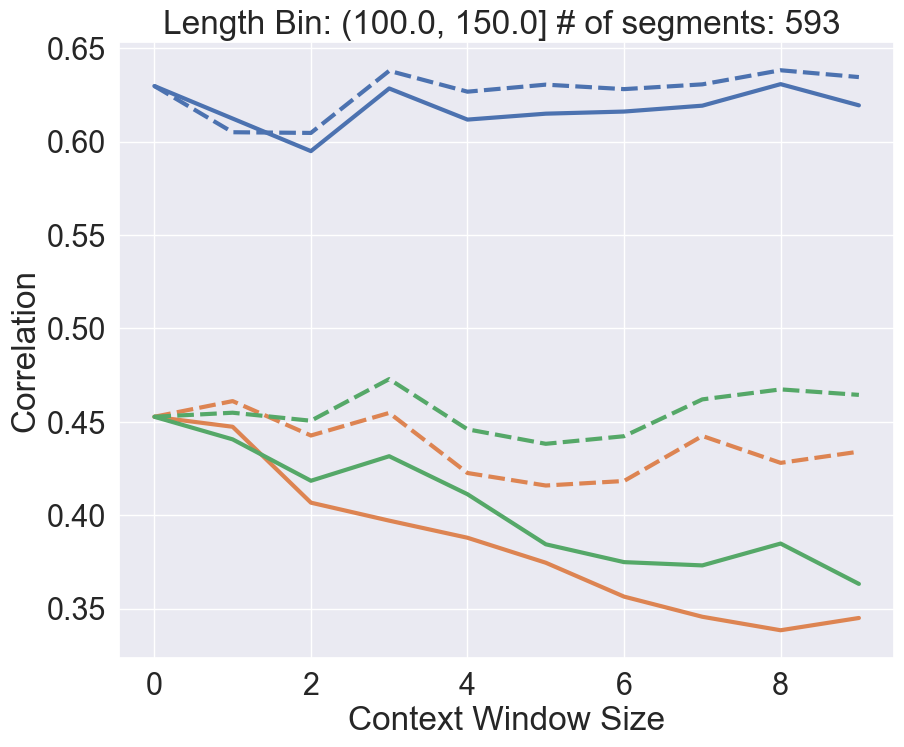}
\end{subfigure}%
    \caption{Context helps the most in improving the translation quality estimation of shorter (source character length $\leq$ 20) and potentially ambiguous sentences (averaged over ``all'' Agent language pairs).}  \label{fig:contextLength}
\end{figure*}

Neural reference-based metrics consistently outperform lexical metrics in most settings, specifically when evaluating translations into English. However, when assessing the translation quality in out-of-English language pairs (Agent), reference-based lexical metrics achieve better correlations with human judgments than neural learned metrics, suggesting room for improvement for reference-free evaluation for assessing translations in languages other than English.

\subsection{Context-Aware Translation Evaluation} \label{sec:contextqe}

We consider the context-aware extensions of two metrics: reference-based 
\textsc{Comet-22} and reference-free \textsc{Comet-20-QE}.
For each of these metrics, we study the impact of adding contextual information as detailed in \S~\ref{sec:chatcontext} in both \textit{within} or \textit{across} participants settings. For reference-free \textsc{Comet-20-QE}, we additionally consider the setup where we use the machine-translated hypothesis instead of the reference as context. 
This is to mimic the real-word chat scenario where references are generally unavailable. We hypothesize that noisy context can still provide useful information in estimating the quality of the current (source, translation) pair.

\subsubsection{Main Results}

Figure~\ref{fig:context} shows results of adding up to last nine sentences as context to the above configurations averaged across customer (``Average Customer'') and agent (``Average Agent'') language pairs. 

\paragraph{Context is not helpful when references are available.} Reference-based \textsc{Comet-22} on average across all language pairs does not benefit (Agent) or hurts (Customer) correlation with the added context information. This could be because most of the necessary information to resolve any ambiguity is ideally already included in the reference. 
Adding more context could potentially introduce ambiguity or inconsistency that is not present in the reference text, hurting the evaluation process.


\paragraph{Adding context is detrimental when evaluating translation into English.} In all settings for customer directions, adding context almost always hurts the metric's correlation with human judgments. Even in the reference-free scenario, translation evaluation into English does not benefit from the added context. We attribute this to the limited contextual phenomena observed in English compared to other languages (PT, FR, DE). 

\paragraph{Adding context improves correlation for \textsc{Comet-20-QE} in reference-free out-of-English settings.} Reference-free \textsc{Comet-20-QE} significantly benefits from the added context on average across all ``Agent'' settings. The correlation increases as the context increases.  Specifically, using the complete contextual information from both participants (\textit{across}) is key to getting the most out of the added contextual information. Shorter segments ($\le 20$ characters) benefit the most from the added context as depicted in Figure~\ref{fig:contextLength}. The above trend holds when using either reference-based or hypothesis-based contexts, which is promising.

\subsubsection{Ablation Analysis}
We use the \textit{across} participant setting with a context window of 2 for \textsc{Comet-20-QE}, as this setup led to the most improvement in correlation. In addition, we use the hypothesis as the context instead of the reference for all the analysis, mimicking the real-world scenario where references are unavailable. With this setup, we first study whether adding context improves correlation for specific error types and severity levels. We then evaluate the impact of adding noise to the context with the proposition that unrelated or partial context should hurt the metric's performance. 

\begin{figure}[t]
\centering
  \includegraphics[width=0.80\linewidth]{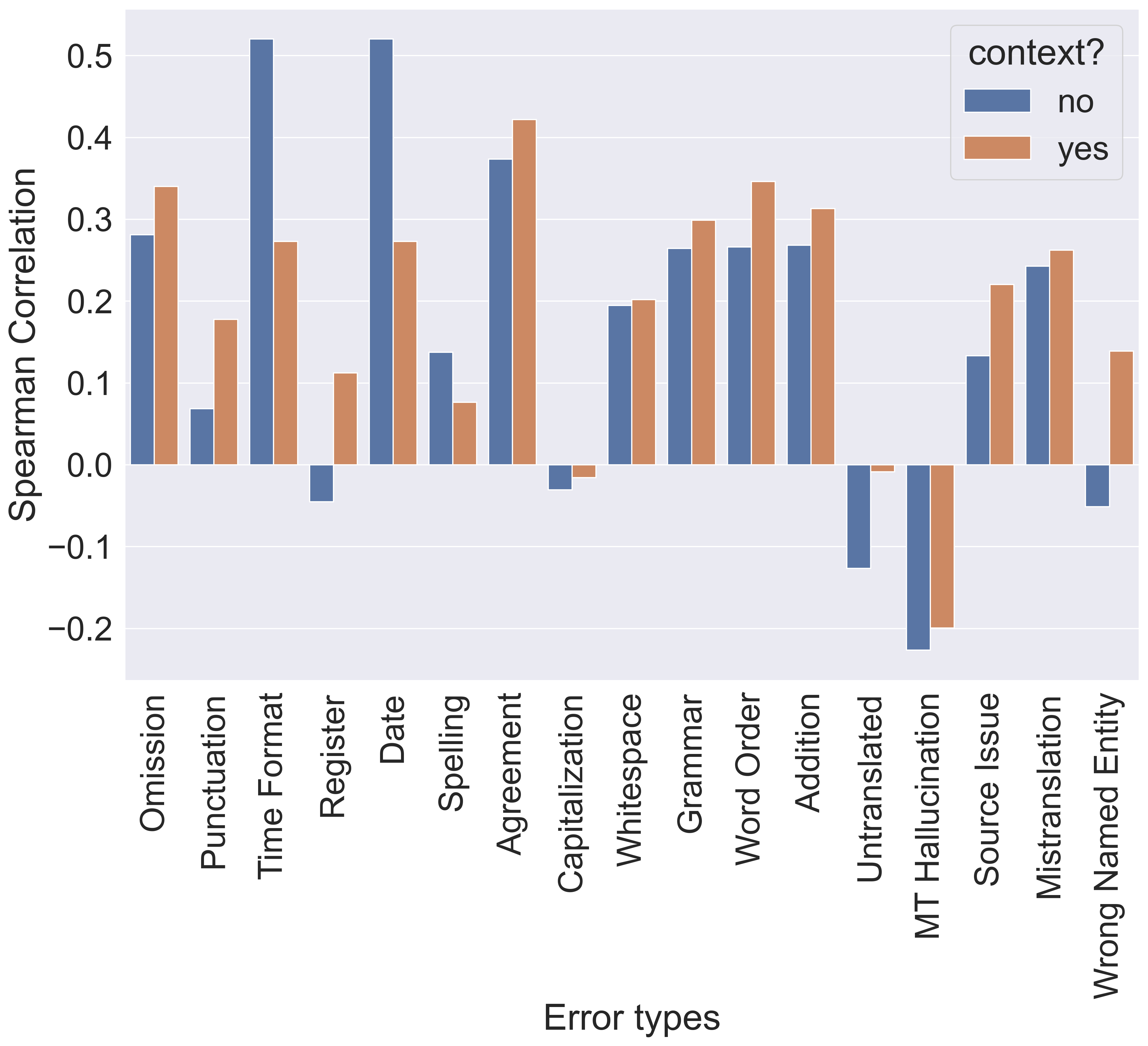}
   \includegraphics[width=0.80\linewidth]{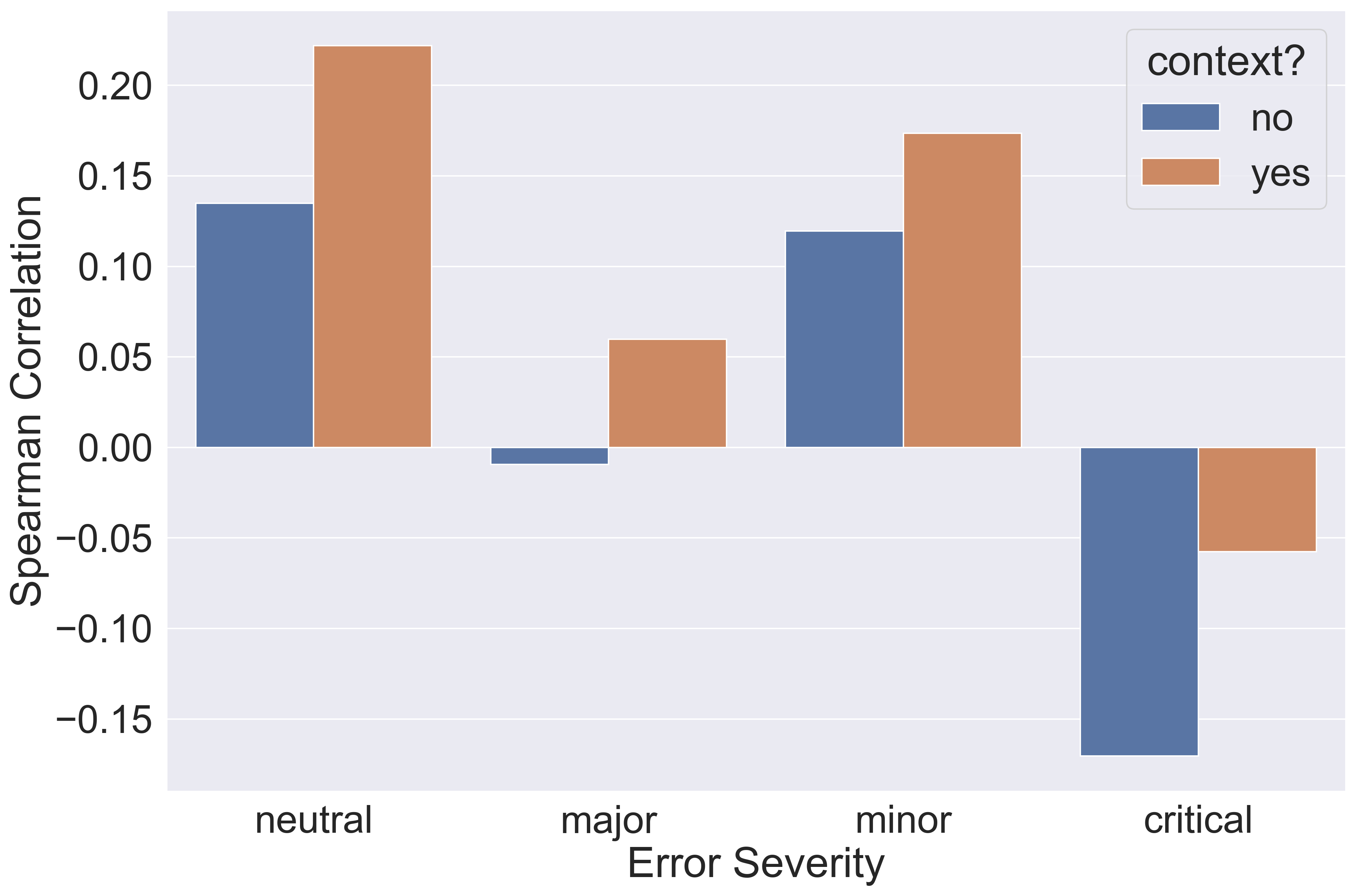}
    \caption{Adding context improves correlation with human judgments across most error types (except Date and Time Format) and all severity levels.}  \label{fig:error_types}
    \vspace{-0.5cm}
\end{figure}

\paragraph{Breakdown by error types and severity} We bucket sentences by the MQM error typology and filter error types with at least 20 instances. Figure~\ref{fig:error_types} shows the correlation with human judgments for each error type bucket and across severity levels. Including contextual information improves correlation with human judgments on several error types as well as all error severity levels. However, adding context hurts Spelling, Date, and Time Format errors, possibly because adding context might introduce noise and not provide any useful information to identify such errors which are solely based on linguistic or formatting rules.   
%

\paragraph{Impact of Noisy or Partial Context}

To validate that complete and correct context is necessary for meaningful improvement in metrics' performance, we inject two types of noise into the context: \textbf{Swap}, where we use unrelated context from a different instance; and \textbf{Drop} where we drop one of the two (source, translation) pairs (``pair'') or unpaired sentences from the preceding context (``random''). We additionally consider injecting these noises into either the source, the translation, or both.

\begin{table}[t]
\centering
\scalebox{0.75}{
\begin{tabular}{llr}
 \toprule
 & \multicolumn{1}{c}{\textsc{Noise to?}} & \multicolumn{1}{c}{\textsc{Avg-Agent}} \\ 
\midrule

\textsc{No Context} & -  & 0.379 \\

\addlinespace[0.2cm]

\textsc{Context} & -  & 0.420 \\
\addlinespace[0.2cm]
\textsc{Swap} & \textsc{Source}  &0.364 \\
&\textsc{Translation}  &0.296  \\
&\textsc{Both}   &0.299 \\
\addlinespace[0.2cm]
\textsc{Drop (Random)} &\textsc{Source} &0.402 \\
&\textsc{Translation}   &0.326 \\
&\textsc{Both}   &0.346  \\
\addlinespace[0.2cm]
\textsc{Drop (Pair)} &\textsc{Source} &0.389\\ 
&\textsc{Translation}   &0.399 \\
&\textsc{Both}   &0.325 \\
  \bottomrule
 \end{tabular}
 }
\caption{Corrupting context hurts correlation.
}\label{tab:noisy_context} 
\end{table}

Table~\ref{tab:noisy_context} shows that adding either kind of noise to the context leads to a drop in correlation relative to the ``Context'' baseline, often even performing worse than using any contextual information (``No Context''). Unrelated context (``Swap'') has a more adverse impact on the metric's performance compared to changing the context via dropping partial information. Furthermore, dropping paired contextual sentences results in a larger drop in correlation than dropping unrelated source-translation instances. This further solidifies our argument that complete and related context is key to utilizing context for chat translation quality estimation.

\begin{figure}[t]
\centering
  \includegraphics[width=0.75\linewidth]{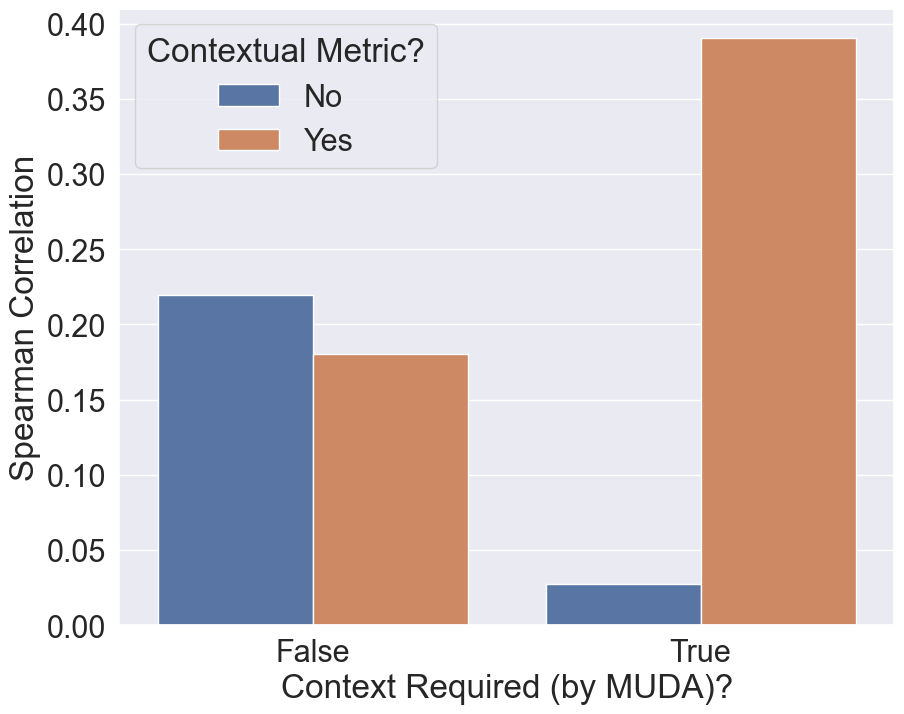}
    \caption{Adding context improves correlation on contextually ambiguous sentences (character length $\le$ 20) for English-German.}  \label{fig:context_muda}
     \vspace{-0.5cm}
\end{figure}

\paragraph{Impact on Contextually Ambiguous Sentences} Given that shorter sentences benefit the most from the added context (Figure~\ref{fig:contextLength}), we further investigate whether this is indeed due to the increased ambiguity in texts. We use \textsc{MuDa} \cite{fernandes-etal-2023-translation} to identify translation pairs with specific discourse phenomena (formality, pronouns, verb form, lexical consistency) for English-German. This enables us to mark instances that potentially require context for disambiguation. Figure~\ref{fig:context_muda} shows that the sentences with marked discourse phenomena benefit the most from the added context. On the other hand, adding context to sentences without marked discourse phenomena hurts correlation. We leave the detailed exploration of when to rely on context for quality estimation to future work.

  \begin{figure*}[htb!]
         \small
         \renewcommand\tabularxcolumn[1]{m{#1}}
        \renewcommand\arraystretch{1.2}
        \centering
    \begin{tabularx}{0.98\linewidth}{*{1}{>{\arraybackslash}X}}
   {\fontfamily{cmss}\selectfont 
  \textbf{System:}  You are an annotator for the quality of machine translation. Your task is to identify errors and assess the quality of the translation.} \\
  {\fontfamily{cmss}\selectfont The categories of errors are: accuracy (addition, mistranslation, omission, untranslated text), fluency (character encoding, grammar, inconsistency, punctuation, register, spelling), style (awkward), terminology (inappropriate for context, inconsistent use), non-translation, other, source error or no-error.}\\
  {\fontfamily{cmss}\selectfont Each error is classified as one of three categories: critical, major, and minor. Critical errors inhibit comprehension of the text. Major errors disrupt the flow, but what the text is trying to say is still understandable. Minor errors are technically errors, but do not disrupt the flow or hinder comprehension.}\\

  \{$k$ Few Shot Incontext Examples\} \\
  {\fontfamily{cmss}\selectfont \textbf{User}:  Context: ```\{context\}'''}\\
   \{sender\} {\fontfamily{cmss}\selectfont source} ({source\_lang}): ```\{source\_seg\}'''\\

    \{target\_lang\} {\fontfamily{cmss}\selectfont translation}:   ```\{target\_seg\}'''\\
   {\fontfamily{cmss}\selectfont Based on the conversation context between the agent and the customer, the current source by "{sender}" in \{source\_lang\} and its machine translation in \{target\_lang\} surrounded with triple backticks, identify error types in the translation and classify them. }\\
   \bottomrule
    \end{tabularx}  
\caption{ Contextual Prompt for Chat Quality Estimation. } \label{fig:prompt_context}
\vspace{-0.4cm}
\end{figure*}

\subsection{LLM-based Contextual Quality Estimation} \label{sec:contextmqm}

Motivated by the recent surge of interest in using LLMs like GPT \cite{openai2023gpt4} for text evaluation, specifically in MT \cite{kocmi-federmann-2023-large, fernandes-etal-2023-devil, lu2023error, kocmi-federmann-2023-gemba}, we also explore their potential in assessing the translation quality of machine translated chats. To better elicit the reasoning and in-context learning capabilities of these LLMs, we prompt GPT-4\footnote{\texttt{gpt-4-0613}, accessed on February 26, 2024.} to identify and categorize errors in machine-generated translations instead of asking for an overall score for translation quality like direct assessment.\footnote{Our initial experiments with open-sourced LLMs suggested limited ability of the models to do well on this task.} 

We present  \textsc{Context-MQM}, 
a context-informed LLM-based quality estimation metric for chat translation evaluation. We adopt the MQM-style prompting techniques to elicit the reasoning capabilities of the LLM \cite{fernandes-etal-2023-devil, kocmi-federmann-2023-gemba} and modify it to utilize contextual information: following Section~\ref{sec:contextqe}, we include the past $k=8$ bilingual source sentences as context and one in-domain in-context example as shown in Figure~\ref{fig:prompt_context}. We perform our evaluation on a subset of 1000 English-German sentences sampled uniformly from the dataset and contrast our metric with the evaluation prompting technique that does not utilize any contextual information, \textsc{LLM-MQM (No Context)}.\footnote{We did not conduct a full-scale evaluation due to the high cost of accessing the GPT-4 API. The cost for running the experiments on English-German was $\sim$ \$300.}

The results are presented in Table~\ref{tab:llm_context}: adding context positively impacts correlation for LLM-based evaluation of machine-translated chats. \textsc{Context-MQM} improves correlation with human judgments, outperforming both non-contextual LLM-MQM (All: +0.013, Imperfect: 0.048) as well as \textsc{Comet-22} (All: +0.091, Imperfect: +0.107). The improvement is larger on the imperfect translations, suggesting that context helps identify errors better on these (source, translation) pairs. 
These initial results show the potential of utilizing LLMs for evaluating chat translation quality with contextual information. Additionally, exploring alternative prompting strategies to integrate context across diverse language pairs and LLMs merits further investigation in future research endeavors.

\begin{table}[t]
\centering
\scalebox{0.8}{
\begin{tabular}{lrr}
 \toprule
 & \multicolumn{1}{c}{All} & \multicolumn{1}{c}{Imperfect} \\
\midrule
\textsc{LLM-MQM (No Context)} & 0.642 & 0.512\\
\textsc{Context-MQM} & 0.655 & 0.560 \\
\midrule
\textsc{Comet-22} & 0.564 & 0.453\\
  \bottomrule
 \end{tabular}
 }
\caption{\textsc{Context-MQM} outperforms \textsc{Comet-22} and non-contextual LLM-MQM on English-German Chat Quality Estimation.
}\label{tab:llm_context} 
\vspace{-0.5cm}
\end{table}



\section{Related Work}


\paragraph{Automatic MT Metrics} Designing automatic metrics to assess translation quality has been an active area of research over the past decade. Metrics shared tasks organized at WMT have significantly facilitated research where recent metrics like BLEURT \cite{sellam2020bleurt} or COMET \cite{rei-etal-2020-unbabels} based on neural architectures and trained with human assessments are shown to consistently outperform lexical metrics. However, these metrics are primarily evaluated by assessing the quality of isolated translations, primarily for news-like data. Recent work, however, has focused on developing document-level evaluation metrics acknowledging that sentences often do not occur in isolation in the wild and the correctness of translation is dependent on the context \cite{voita-etal-2019-good}. Document-level metrics like SliDe \cite{raunak-etal-2023-evaluating} or BlonDe \cite{jiang-etal-2022-blonde} use discourse information to assess the translation quality at the paragraph level. In this work, however, our goal is still to assess the translation quality of individual sentences but inform the quality evaluation with the available contextual information.

\paragraph{Chat Translation Quality Estimation} \citet{li-etal-2022-chat} introduce the erroneous chat translation detection task and propose an error detection model that classifies a given translation in a bilingual \textit{two-utterance} chat as either correct or erroneous. However, their approach requires training a detection model tailored explicitly to chat-related data, whereas, in this work we benchmark existing metrics for general quality estimation and study the impact of conversation context on quality estimation via existing metrics. 

\citet{menezes-etal-2023-context} propose a new framework for identifying contextual errors in conversational datasets. They expand the MQM categories to account for errors introduced due to \textit{contextual triggers}. They further show that these errors are indeed critical and that current metrics fall short in detecting them. Our evaluation instead targets estimating chat translation quality regardless of the specific errors and is aimed at providing a more general view of the ability of existing metrics to assess the quality of machine-translated chats. 

\paragraph{Contextual Machine Translation} In many scenarios, translation requires leveraging information beyond the sentence level to resolve inter-sentence dependencies and improve translation quality. Incorporating context to generate high-quality translations has been explored for conversation and news documents, with approaches ranging from simply concatenating the context to the original input \cite{Tiedemann2017NeuralMT} to more complex options \cite{Jean2017DoesNM, maruf-etal-2018-contextual,maruf-etal-2019-selective}. However, despite having access to context, contextual MT models still struggle to effectively use it \cite{fernandes-etal-2021-measuring}, and most MT metrics fail to capture this due to a lack of utilization of context in the evaluation metrics themselves. 
Hence, we instead use context to assess the translation quality of a given (source, target) pair and show that it benefits the evaluation of non-English translations.

\section{Conclusion and Discussion}

Estimating translation quality in diverse domains is crucial to ensure that the metrics employed in MT evaluation accurately reflect the MT system's quality across various types of content. We show that the nature and the type of errors in the conversational context are different from the generally evaluated news domain. Hence, designing robust metrics that can capture these errors is very important. Our work presents a step in that direction by systematically benchmarking existing automatic MT metrics on machine-translated chats. Given the highly contextual nature of the chat domain, we extend and evaluate context-based reference-free and reference-based metrics and show that context can be extremely helpful in judging translation quality, especially in a reference-free setup. Our analysis sheds some light on how and when the context can be helpful. However, there remain several open questions and directions for future work.

\paragraph{Improving Detection of MT errors} As illustrated by our results, although \textsc{Comet-22} achieves high correlations with human judgments overall, there is a drop in correlation for imperfect segments, suggesting a need for designing a metric that can do well at estimating quality for both perfect and imperfect translations.

\paragraph{Better Reference-free Evaluation} Our findings show that reference-free learned metrics lag behind reference-based ones in evaluating the translation quality of bilingual chats. This presents opportunities to develop effective evaluation methods in the absence of reference translations.

\paragraph{Optimizing Context Utilization} We implemented a very simple approach to utilize context from both participants in both learned metrics as well as for LLM-based MT evaluation. However, it remains to be investigated how context can be utilized in other ways and when the metric should rely on the contextual information.

\section*{Acknowledgments}

We thank Ben Peters, António Farinhas, and Duarte Alves for their useful and constructive comments. This work was supported 
by the Portuguese Recovery and Resilience Plan through project C645008882-00000055 (Center for Responsible AI), by the EU’s Horizon Europe Research and Innovation Actions (UTTER, contract 101070631), by the project DECOLLAGE (ERC-2022-CoG 101088763), and by Fundação para a Ciência e Tecnologia through contract UIDB/50008/2020.

\bibliography{tacl2021}
\bibliographystyle{acl_natbib}


\end{document}